\documentclass[a4paper,fleqn]{cas-dc}

\usepackage[numbers]{natbib}
\usepackage{comment}
\newcommand\nnfootnote[1]{%
  \begin{NoHyper}
  \renewcommand\thefootnote{}\footnote{#1}%
  \addtocounter{footnote}{-1}%
  \end{NoHyper}
}

\def\tsc#1{\csdef{#1}{\textsc{\lowercase{#1}}\xspace}}
\tsc{WGM}
\tsc{QE}

\usepackage{calc}
\usepackage{amssymb}
\usepackage{amstext}
\usepackage{amsthm}
\usepackage{mathtools}
\usepackage{amsmath}
\usepackage{multicol}
\usepackage{plain}
\usepackage{tabularx}
\usepackage{float}
\usepackage{systeme}
\usepackage[T1]{fontenc}
\usepackage{fancyhdr}
\usepackage{algorithm}
\usepackage{algpseudocode}
\usepackage{diagbox}
\usepackage{multirow}
\usepackage{hyperref}
\newcolumntype{M}[1]{>{\centering\arraybackslash}m{#1}}
\usepackage{graphicx}
\usepackage{subfigure}
\usepackage[font=footnotesize,labelfont=bf]{caption}
\usepackage{xparse}

\newcommand{\Black}[1]{\textcolor{black}{#1}}

\makeatletter
\NewDocumentCommand{\raisedminus}{m}{%
  \raisebox{0.2em}{$\m@th#1{-}$}%
}

\makeatother
\begin{document}

\nnfootnote{.}
\def\floatpagepagefraction{1}
\def\textpagefraction{.001}

\shorttitle{Online motion recognition}    

\shortauthors{M.S.AKREMI, R.SLAMA, H.TABIA}  

\title[mode = title]{Accurate online action and gesture recognition system using detectors and Deep SPD Siamese Networks}  

\tnotemark[1] 


%

\author[1]{Mohamed Sanim AKREMI}[orcid=0000-0000-0000-0000] 



\ead{mohamedsanim.akremi@univ-evry.fr}


\credit{<Credit authorship details>}

\affiliation[1]{organization={IBISC, Univ Evry. Université Paris Saclay },
            city={Evry},
            state={Paris},
            country={France}}

\author[2]{Rim SLAMA}[style=arabe]


\ead{rim.slamasalmi@entpe.fr}
\affiliation[2]{organization={Univ. Eiffel}, addressline={ENTPE, LICIT-ECO7}, postcode={F-69518}, city={Lyon}, country={France}}  





\author[1]{Hedi TABIA}[style=arabe]


\ead{hedi.tabia@univ-evry.fr}







\begin{highlights}
\item The "SPD Siamese Network" is efficient to resolve motion recognition problem
\item An online system based on a detector and a classifier achieves well performances 
\item The proposed algorithm is accurate in different contexts and on challenging datasets 
\end{highlights}

\begin{abstract}
Online continuous motion recognition is a hot topic of research since it is more practical  in real life application cases.  Recently, Skeleton-based approaches have become increasingly popular, demonstrating the power of using such 3D temporal data. However, most of these works have focused on segment-based recognition and are not suitable for the online  scenarios. In this paper, we propose an online recognition system for skeleton sequence streaming composed from two main components: a detector and a classifier, which use a Semi-Positive Definite (SPD) matrix representation and a Siamese network. The powerful statistical representations for the skeletal data given by the SPD matrices and the learning of their semantic similarity by the Siamese network enable the detector to predict time intervals of the motions throughout an unsegmented sequence. In addition, they ensure the classifier capability to recognize the motion in each predicted interval. The proposed detector is flexible and able to identify the kinetic state continuously. We conduct extensive experiments on both hand gesture and body action recognition benchmarks to prove the accuracy of our online recognition system which in most cases outperforms state-of-the-art performances.
\end{abstract}
\maketitle
\section{Introduction}\label{Introduction}
Human activity recognition is an important research topic in pattern recognition field. It has been the subject of many studies in the past two decades because of its importance in numerous areas such as security, health, daily activity, energy consumption and robotics.
Recently, some works on the recognition of hand gestures or human actions from skeletal data are based on the modeling of the skeleton’s movement as manifold-based representation and proposed deep neural networks on this structure ~\cite{huang2018building,huang2017riemannian,vemulapalli2014human}. These approaches demonstrated their potential in the processing of skeletal data. Most of them are applied on offline human action recognition which is useful in time-limited tasks. 
However, in many applications, simply recognizing a single gesture in a given segmented sequence is not enough, especially in monitoring systems and virtual-reality devices which need to detect human movements moment by moment in continuous videos.

In these online recognition systems, it is important to detect the existence of an action as early as possible after its beginning. It is also essential to determine the nature of the movement within a sequence of frames, without having information about the number of gestures present within the video, their starting times or their durations, unlike the segmented action recognition.

In this paper, we propose to use a manifold-based model in order to build an online motion recognition system that detects and identifies different human activities in unsegmented skeletal sequences. 
As demonstrated in our previous work ~\cite{visapp22}, the SPD Siamese network is accurate for hand gesture classification. Thus, we decide to generalize our work on 3D body skeleton representation and adapt it for action recognition problem. 
Besides, we upgrade our model to make it as suitable as possible for online recognition criteria ~\cite{kopuklu2019real}. 

Our proposed online recognition system is composed from (i) a detector which segments the sequence and (ii) a classifier which identifies the action or the gesture.

In the following, we review both segmented and continuous human motion recognition methods. A focus is given to manifold deep learning approaches applied to 3D skeleton data in the offline recognition approaches. For the Online recognition system, we review the approaches applied for sequence segmentation and online recognition, whatever the type of data. An overview of our approach is also given to outline the main steps of the proposed solution.
The pretrained models and the code are provided at the following \href{https://github.com/Mohamed-Sanim/Online-motion-recognition}{link}.
\subsection{Related works}
Many methods based on manifold learning approaches are proposed in proceeding 3D skeletal data. 
Vemulapalli et al. ~\cite{vemulapalli2014human} worked on the special Euclidean Group denoted by SE(3). They mapped the action curves from SE(3) to the lie Algebra se(3). Then, they computed a nominal curve using Dynamic Time Warping (DTW) and used Fourier Temporal Pyramid (FTP) representation to eliminate the noise issue before the use of the classification with SVM.  Huang et al. ~\cite{huang2017deep} proposed a Lie group Network (LieNet) which based on an input rotation matrix, applies mapping transformation in order to construct the best rotation matrix.
Other studies focus on the SPD manifold-based approaches: a  Riemannian metric learning for SPD Matrices was proposed by Huang et al. ~\cite{huang2017riemannian}; a learning matrix neural network using mean and covariance statistics was proposed by  ~\cite{nguyen2019neural} .
A deep neural network on Grassmann manifold, denoted by GrNet, was proposed by ~\cite{huang2018building}. It is composed of 3 major blocks: a projection block used for the transformation of the orthonormal input matrices, a pooling block designed to map the orthonormal matrices and apply a mean pooling on them, and an output block utilized for mapping and classification.
 
These approaches, as proposed, work well on segmented sequences. In order to adapt them to an online recognition task, introducing online approaches into these methods is required.
The well-known adopted strategy in identifying online motions is based mainly on determining the time interval for each motion, then brings the problem to an offline recognition task. Some studies, such as ~\cite{devanne2019recognition}, went to monitor movements by studying the changes that occur during the transition from stagnation to active state. Then, they analyzed spatially limbs behavior in the stagnation. However, this automatic segmentation did not generally obtain the required efficacy. Some works have also gone away from segmentation and are content with knowing the movement locally ~\cite{kopuklu2019real,li2016online}. Other approaches proposed various sliding window strategies for online recognition purpose \cite{delamare2021graph,negin2018online, caputo2020sfinge}.
A Temporal Recurrent Network (TRN) on fixed length windows was proposed by ~\cite{xu2019temporal}  to anticipate the future using a temporal decoder incorporating  future information with the historical information to improve online action detection. Delamare et al.~\cite{delamare2021graph} used a sliding window approach to execute Spatio-Temporal Graph Convolutional Network (ST-GCN).
\subsection{Contributions and method overview}
The major contributions of this paper are as follows: (1) \Black{A new portioning for the human body skeleton  is proposed, enabling the "SPD Siamese Network" to operate on action recognition context. (2) A detector and a verification process are introduced in order to upgrade our method from an offline  to an online system able to identify hand gestures or body actions in unsegmented sequences with high performance and fast reaction.} (3) The efficiency of the proposed algorithm is proved in different  contexts and datasets.

First, we propose to build a network that recognizes hand gestures and body actions using a proposed partitioning of body/hand joints, SPD matrices spatio-temporal representations of the skeletal sequence and the characteristics of the Siamese network.

\begin{figure}[t]  
    \centering
    \includegraphics[width=0.48\textwidth]{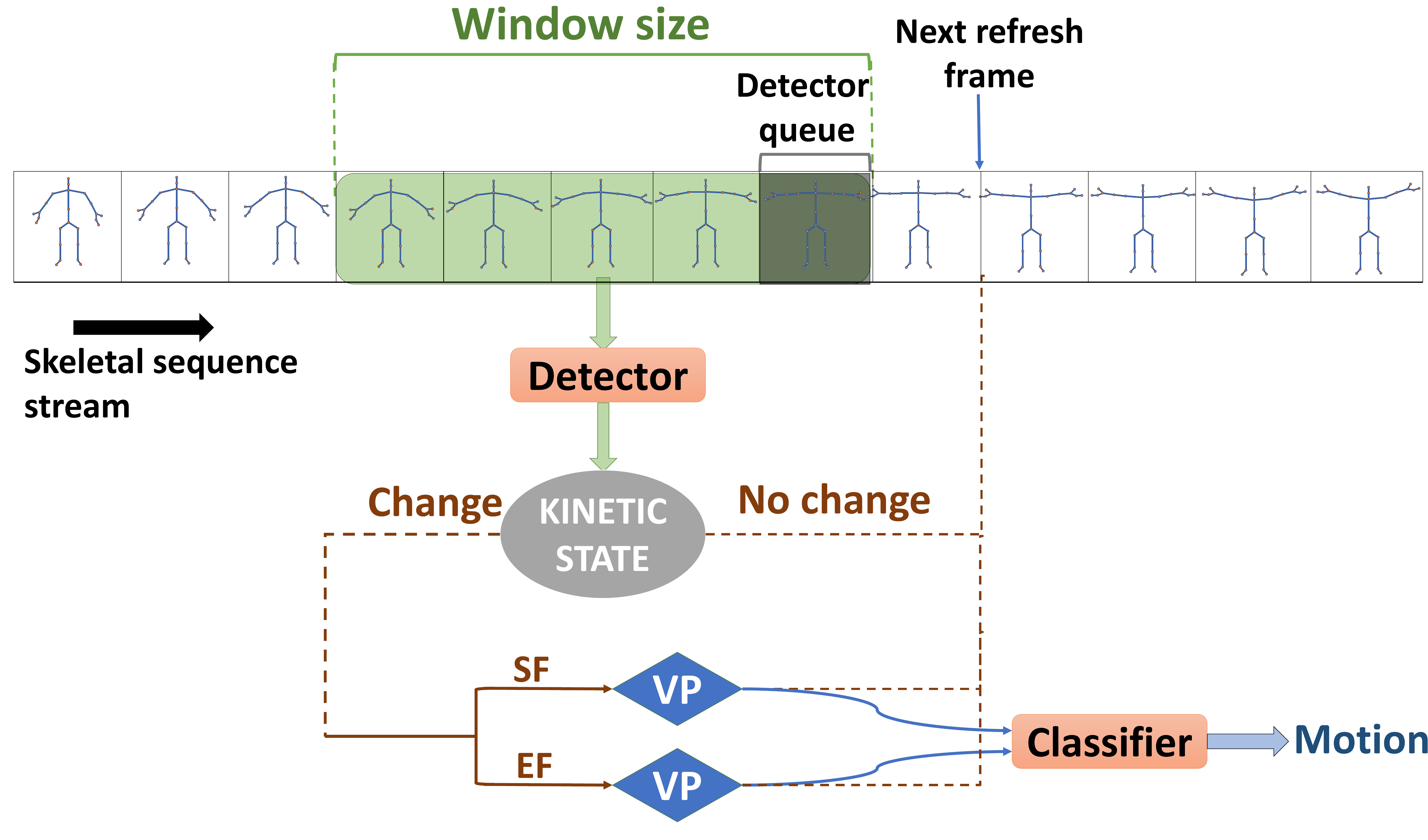}
    \caption{The proposed online motion recognition system. 
    }
    \label{architecture}
\end{figure}

In a next stage, we build an online system based on two main components: a detector and a classifier that use the proposed SPD Siamese network for the identification of the kinetic state and the recognition of the motion respectively.
The detector has as a role to identify continuously the kinetic state and find the segments of each performed motion throughout a long sequence: it predicts the starting frame and the end frame of each performed motion by detecting a change in the kinetic state of the volunteer. As illustrated in Figure ~\ref{architecture}, the change in the kinetic state may refer to an action beginning, an action end or an unexpected idle period. In order to verify the nature of the state, we realize a Verification Process (VP). After each segmentation,  the classifier is activated in order to recognize the motion segment.

The rest of the paper is constructed as follows: section 2 presents the proposed network for motion recognition. Section 3 presents the aspects of our online system. In section 4, the experiments results are reported.  Finally, the last section is dedicated for the conclusion.

\section{SPD Siamese neural network for motion recognition} \label{spd-siamese}
The SPD Siamese network aims to build a classifier capable of recognizing the actions performed by humans, and  a detector that identifies the kinetic state. It uses specific partitioning for the body and hand joints. Then, it learns an SPD matrix representing a motion sequence. Finally, a Siamese network is used to finalize the classification of this sequence. 
\subsection{Proposed architecture}
\begin{figure}[t]
  \centering
  \includegraphics[width=0.45\textwidth]{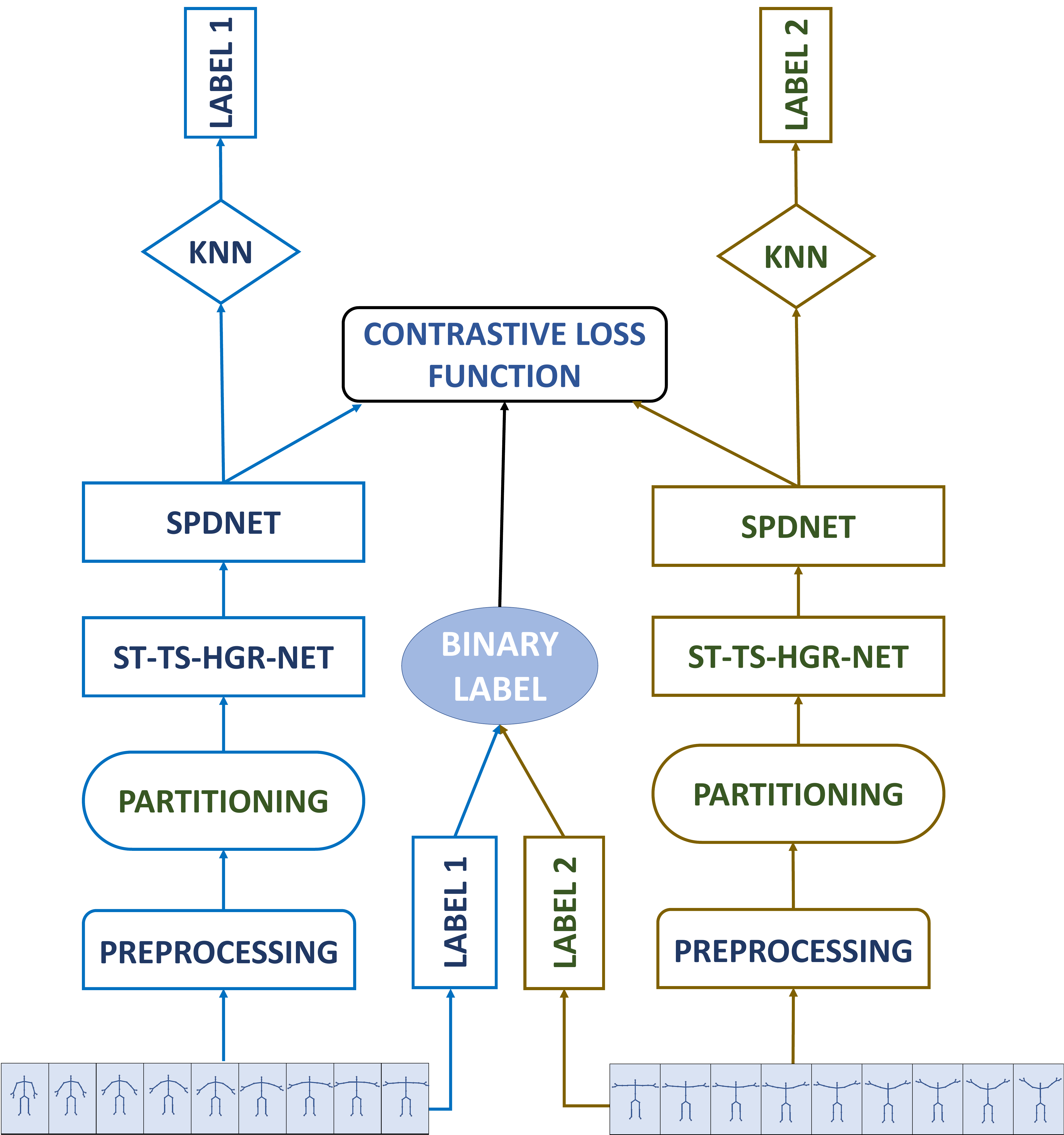}
  \caption{The overview of the proposed SPD Siamese network.}
  \label{model}
\end{figure}

The proposed network, illustrated in Figure ~\ref{model}, is composed of five principal components: preprocessing component, partitioning component, SPD learning features component, SPD Siamese network component, and classification component. 

Having the 3D skeleton sequence, we divide the joint set into parts in order to perform the spatial analysis. The joints from each part must have a correlation between each other.
In the SPD learning component, we analyze the spatial-temporal evolution and the temporal-spatial evolution in order to obtain the best SPD matrix representing the skeleton sequence.
For the SPD Siamese network component, we use as a base model the network proposed by ~\cite{huang2017riemannian} without transformation blocks since we realized the SPD learning in the previous component.
We twin the two previous components and use the contrastive loss function to train our model. Finally, we use the K-NN algorithm on the learnt model parameters applied to the base SPD network component for the classification.

\subsection{Proposed representation for body/hand parts} \label{PARTS}
\begin{figure}[ht]
    \centering 
    \includegraphics[width=0.48\textwidth]{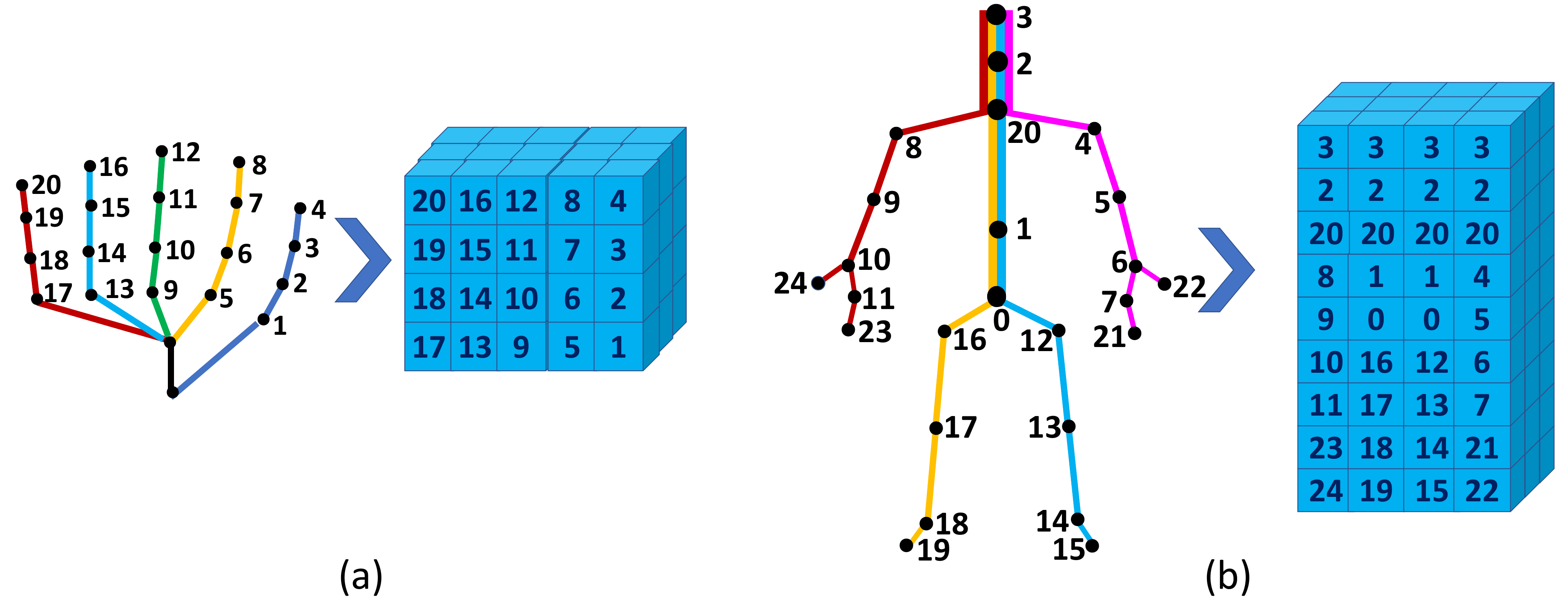}
    \caption{Skeleton parts and matrix representation for (a) the hand and (b) the body.}
  \label{handmr}
\end{figure}

Having 3D skeletal joint coordinates, we divide the hand/body skeleton into parts in order to study the evolution of the coordinates locally. \Black{As explained in ~\cite{visapp22}}, for the hand, we consider that each finger represents a part (Figure ~\ref{handmr}(a)). As for the body, \Black{the proposed partitioning consists in combining each joint of the body to an adjacent joint, starting from the head down to the other parts of the body. When a joint has more than one adjacent, it produces ramifications which we follow and divide the body into  into four parts (Figure ~\ref{handmr}(b))}: 
 (i) upper right part that runs from the head down to the palm of the right hand,
(ii) upper left part that goes from the head part to the palm of the left hand,
(iii) lower right part which includes the spine connected to the right leg,
(iv) and lower left part which includes the spine connected to the left leg.
\subsection{SPD matrix learning}
In order to learn the better SPD matrix representation, we use the ST-TS-HGR-Network inspired by ~\cite{nguyen2019neural}.
This network is composed from four principal phases. The first phase is the convolution layer. It highlights the correlation between the neighboring joints, and it learns the filter weight associated to each neighbor of a given joint. \Black{We apply 3×3 convolution on the matrix representation of the partitioning.}
Then, we divide the sequence into 6 sub-sequences\Black{: the first represents the whole sequence, the second and the third represent the two halves of the sequence and the rest represents its three thirds.} We perform a spatial-temporal analysis in the ST-GA-NET phase and a temporal-spatial analysis TS-GA-NET phase of body/hand parts evolution. \Black{Each phase contains five layers. The first layer, Gaussian Aggregation (GA) outputs the SPD matrix $Y = 
\begin{bmatrix}
\Sigma + \mu \mu^{T} &  \mu \\ 
 \mu^{T} &  1 
\end{bmatrix} $, where $\mu$  and $\Sigma$ denote respectively the mean and the covariance of the joints positions spatially in ST-GA-NET and temporally in TS-GA-NET. Then, we apply a ReEig layer to rectify the SPD matrices, outputs of GA layer, by tuning up their small positive eigenvalues using a chosen threshold. The LogEig and VecMap layers are used to map the resulting SPD matrices into the euclidean space. Then, a second GA layer computes the mean and the covariance along the temporal domain in ST-GA-NET and along the spatial domain (parts) in TS-GA-NET}.
In the final phase, SPDC-NET  has as a function  to generate more compact and discriminative SPD matrix. \Black{It takes as input the resulting SPD matrices from both ST-GA-NET and TS-GA-NET, returning as output $Y =  \sum W_{i}X_{i}W_{i}^{T}$, where its parameters $\{W_{i}\}$  are Stiefel weights and $\{X_i\}$ are the outputs from both previous components. For more details on equations used in each layer, we refer readers to ~\cite{visapp22}}.
\begin{figure*}[b]
  \centering
  \includegraphics[width=0.95\textwidth]{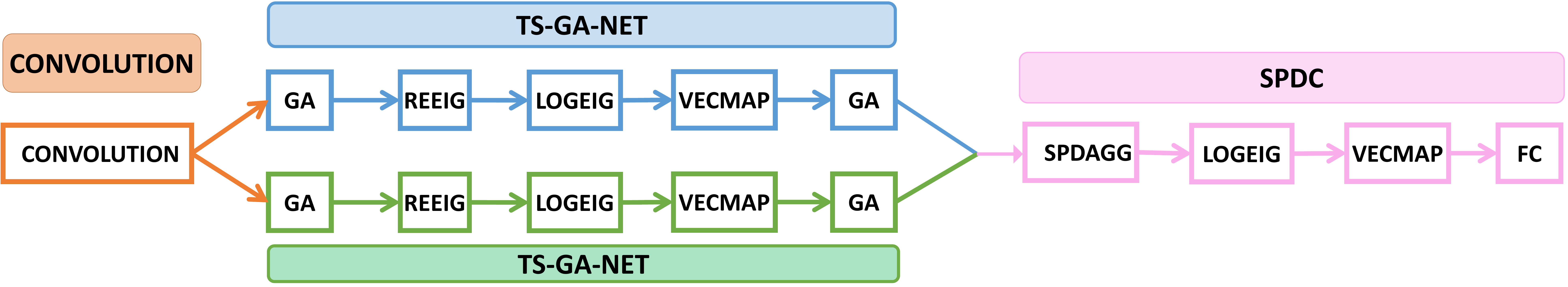}
  \caption{ST-TS-HGR-NET architecure}
  \label{st-ts}
\end{figure*}
\subsection{SPD Siamese network and motion recognition}
\label{SPD Siamese network}

The SPD Siamese network consists of two identical SPD sub-networks joined at their outputs and it is characterized by its margin parameter $g$. We have as input, pair of two SPD matrices, with a binary label $b$. In each sub-network of the SPD Siamese network, we map an SPD matrix into a tangent space, and we extract a feature vector using a fully connected layer. Then, we use the Contrastive Loss ($CL$) function to measure the distance between the two extracted feature vectors and minimize distance between positive pairs using the following equation: ${Cl(y_1,y_2,b) = b||y_1 - y_2 ||_2 + (1 - b) max(0, g-||y_1 - y_2||_2)}$
Where $y_1$ and $y_2$ are the outputs of the two twin SPD sub-networks and ||.||$_2$ is the Euclidean distance.
For the motion recognition, we apply K-Nearest Neighbor (K-NN) algorithm with $K =1$.
\section{Online action and gesture recognition using a detector}
\label{online}
The effectiveness of our online system is related to the performance of the proposed detector and its ability to predict the motions intervals throughout a long sequence. We are also interested to study the efficiency of our system within an early classification.
\subsection{Proposed detector}
The detector preparation goes through two stages: training loop and testing loop. As a first step, we train the detector with the segmented sequences. Then, we evaluate its performance on unsegmented sequences. 

\textbf{Training loop:}
In most sequences, there is a period of inactivity between every two successive motions, called idle period or stagnation state. \Black{In order to distinguish between such an idle period (State "0") and a period of activity (State "1") and detect the transition of state, we use a binary detector.}

Sometimes, we have sequences in which there is not necessarily an idle separating between two actions.
\Black{In this case, we use a multi-class object detector to  detect the transition of the kinetic states between one action and another. Let N be the number of classes in the training dataset. While a subject is performing an action $i$, we attribute the state $i$ to each possible subsequence extracted from its interval. When a transition from a state $i$ to another state $j$ is detected and verified, we consider that there is a transition from an action to another.}
Actually, in the training process, we extract from each sequence randomly subsequences with equal size: $ws$ frames. This size should be long enough for a person to perform part from an action (longer than half a second). We attribute to the subsequence the dominant state class, i.e., when the trained sample contains two different kinetic states, it will be labeled according to the longer state within it.
\Black{Same as the classifier, we use the SPD Siamese Network to train the detector (see section~\ref{spd-siamese}).}

\textbf{Testing loop: }
\Black{In the testing loop, we adopt the sliding window method proposed by ~\cite{delamare2021graph}, augmenting it by the trained detector and a Verification Process (V.P) that helps our system in reducing False Positive (FP) rate and giving more accurate segmentation. Unlike the method proposed by ~\cite{negin2018online} who relies on clustering sub-activities using his proposed detector (that he calls SSD), we propose a supervised method to train our detector. Instead of having unknown numerous groups of sub-activities, we consider these sub-activities belonging to more general classes ("0" and "1" in the case of a binary detector). Besides, we use majority voting rather than Markov probability matrix for sequence segmentation. since in case of high similarity of sub-activities (like in "Swipe X" and "Swipe V"), the performance of Markov probability matrix in predicting the classes of the consecutive sub-activities is generally not accurate.}

Let $ws$ be the size of the window (the same size of the training subsequences), $r$ be the refresh rate of the detector and $cr$ be the capturing rate of the used sensor.
Each $r$ frames, the detector reveals the kinetic state on each window, i.e., the segments on which the detector is applied are $[0, ws]$, $[r, ws + r]$, $[2r, ws + 2r]$... As we are dealing with continuous sequences, we have to set $r\le 0.3 * cr$ frames and ensure that the running time of the detector does not exceed $r$ frames.

\begin{figure}[ht]
  \centering
  \includegraphics[width=0.48\textwidth]{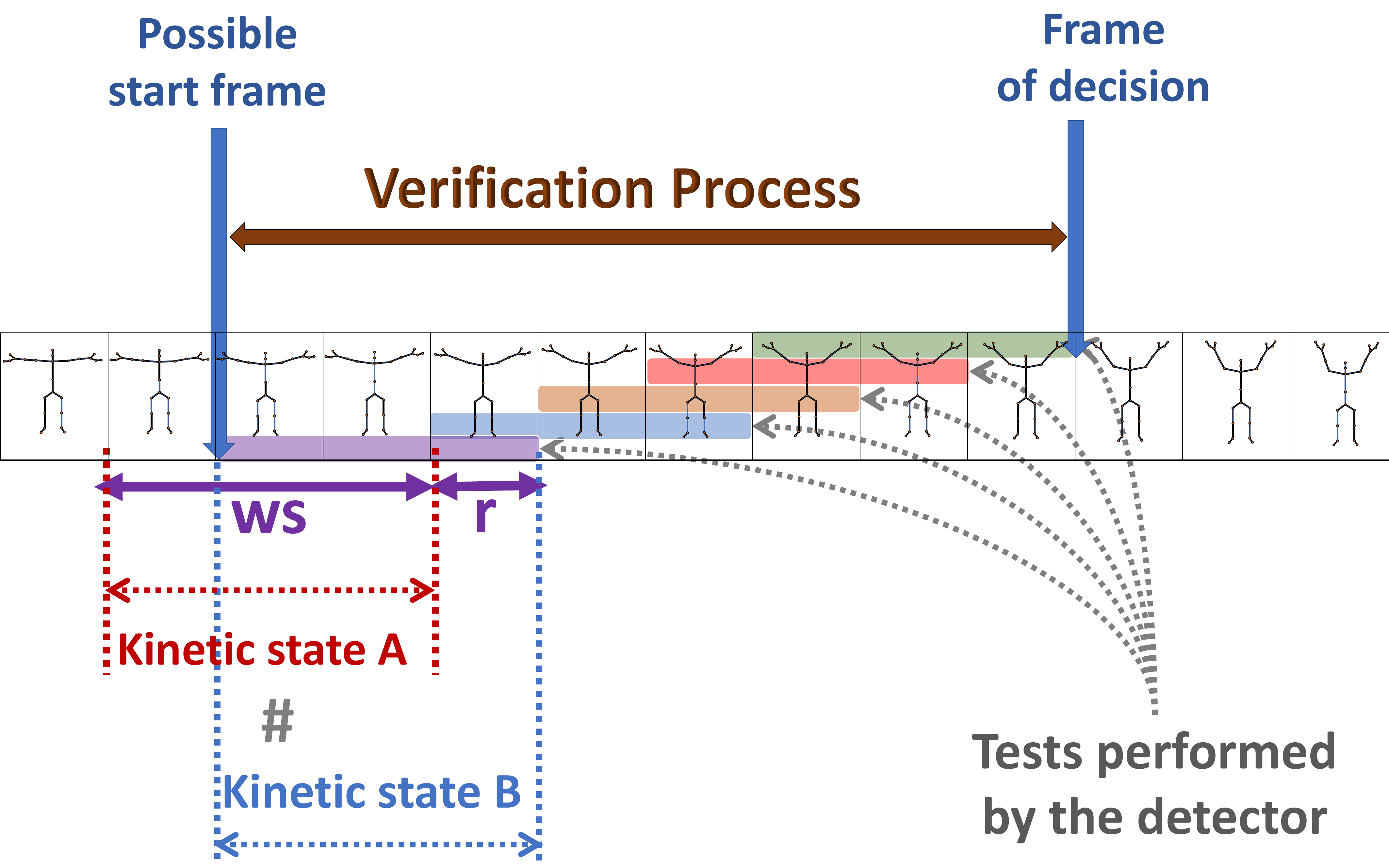}
  \caption{Verification process with $te=5$. }
  \label{verif}
\end{figure}

\Black{When the state of the detector moved from a state $A$ to a different state $B$, we need to verify if it is a real change of a kinetic state or a false detection. To this end, we realize a verification process based on a bunch of repetitive tests to confirm this transition. Let $te$ be the number of these tests. The verification starts from the second test until the $te^{th}$ test while retaining the states of the detector within each window. We use the method of "majority voting" on states resulting from each test to make our decision on the transition to be checked.}
For example (see Figure \ref{verif}), with $te = 5$, when a binary detector detects a transition in the kinetic state from  state "0" to "1" at the window ${[N- ws,N]}$, we realize 4 other tests on the segments ${[N + k \times r -ws, N + K \times r]_{(k=1,2,3,4)}}$. 
We suppose that the states detected are $[0, 1, 1, 1, 0]$. The change of state is confirmed in 3 tests among 5. So, we can conclude that $(N - r)^{th}$ frame is the starting frame of an action.
It is important to notice that the detected starting frame is recognized only after  ($te$ × $r$) frames. So, we need to  minimize $te$ and $r$ as much as possible, relying on a high performing detector with fast reaction time.
\subsection{Early classification}
\Black{In the context of daily human activities ~\cite{li2016online,tang2015online} or hand gesture context ~\cite{Smedt_2016_CVPR_Workshops,caputo2021shrec}, we can tolerate a delay of no more than one second. However, in industrial context ~\cite{9209531} no delay is tolerated, the online recognition should be instantaneously even before the action is finished (Early classification is needed (see Figure~\ref{early-class}).
}
We define $T$ as the maximum time used to recognize a motion.
In order to ensure that we will not exceed $T$ seconds, we set $te \le \frac{T}{r} * cr$. 
Once a frame is recognized as the starting frame of a motion, we wait until the next $T$  frames coordinates are provided. Then the classifier recognizes the motion. If the end frame of a motion is detected before reaching the deadline we set, the classifier will give the class of the motion between the two predicted boundaries, just like the no-early classification case.

\section{Experiments and results}
Our method works for both online  hand and action recognition purpose.
Thus, we evaluate it on four challenging datasets captured in different contexts.
In the following, we extensively  provide details on the experimental settings and results obtained for each dataset. We also  compare the state-of-the-art methods to our approach using  online metrics. Our model is implemented on an octa-core CPU running at 3.2 GHz  with 32GB RAM in python 3.9.7 environment.
In the whole sequence, R.T denotes running time.
\subsection{Datasets}
\textbf{Online Dynamic Hand Gesture dataset (ODHG)} ~\cite{Smedt_2016_CVPR_Workshops}: It consists of 280 relatively long video clips taken with 28 volunteers. Each video contains 10 hand gestures. The data is captured by the depth camera (30 fps) and consists of 14 hand gesture sequences performed in two ways: using one finger and the whole hand.

\Black{\textbf{SHREC 2021 gesture benchmark dataset} ~\cite{caputo2021shrec}: It consists of 180 unsegmented hand skeletal sequences captured by LeapMotion sensor at rate of 25 fps. 60\% of these sequences are used in the training loop and and the rest are dedicated for the test.
The dataset dictionary is made of 17 gestures divided in static gestures and dynamic gestures.}

\textbf{Online Action Detection dataset (OAD)} ~\cite{li2016online}: It includes 59 long sequences representing 839 actions, captured at 8 fps. The possible actions revolve around 10 action classes. It gives the coordinates of 25 joints from different body parts. We have 30 sequences for training and 20 sequences for testing.

\textbf{UOW Online Action 3D dataset} ~\cite{tang2015online}: 
It consists of  48 skeleton sequences recorded using the kinect V2 sensor (20 fps). It has 21 action classes.
Without a stagnation between two consecutive actions, its actions are performed with two manners: repeatably and continuously. 
The sequences containing repeated actions are used in the training and the continuous sequence is dedicated for the testing loop. 

\textbf{Industrial Human Action Recognition Dataset (InHard)} ~\cite{9209531}: It is collected in industrial context of human robot collaboration. It contains 38 long sequences: 26 sequences used in the training and 12 sequences are dedicated for the test. The possible actions are 13 Meta actions. The skeletal data comprises the 3D coordinates and the 3 rotations around the axis of 21 body joints.

\subsection{Evaluation of the classifier on segmented sequences}
The experiments performed for the evaluation of the classifier follow the concept of the segmented motion recognition. The training process and the validation of the classification are based on the proposed SPD Siamese network.
First of all, we start by data cleaning to remove corrupt or inaccurate records.
For InHard dataset, we remove two long sequences P04\_R01 and P04\_R02 in which there is no description of the evolution of the 3D coordinates.
Moreover, for this dataset, we remark that working with the derivates of the sequences gives better performance comparing to that given by the position evolution studying.

After data preprocessing, we prepare the segmented sequences. Using the groundtruth information, we extract the actions sequences performed within the continuous sequences with respect to their train/test protocols mentioned in their description.
In the partitioning component (as illustrated in Figure ~\ref{handmr}), we follow the matrix representation of the hand parts for ODHG and \Black{SHREC 2021} datasets and the matrix presentation of the body parts for the other datasets.

As preprocessing, the input sequences are normalized and interpolated to 500 frames for ODHG and \Black{SHREC 2021}, to 200 frames for OAD and UOW  datasets and to 600 frames for InHard. The difference of the number of the interpolated frames are due to the difference between the capturing rate and the average duration of the motion in each dataset. For the rest of the components, we keep the same configurations set on the experiment section of ~\cite{visapp22}.

\Black{\textbf{Body partitioning strategies:}
Different partitioning strategies are assessed and compared in order to select the more efficient one for body skeleton. As illustrated in Table ~\ref{partitioning}, three strategies are considered. The first proposed by ~\cite{yekini2016market}, divide the skeleton into 5 parts, the second, proposed by Ji et al. ~\cite{ji2018skeleton} divide the body into 10 parts and the third is the one we proposed. Experiments, conducted on segmented sequences of OAD dataset have proven the efficiency of the third strategy we propose in term of classification accuracy. 
This can be explained by the fact that our strategy created more interconnection between adjacent joints and enhanced it in the convolution component. In the following, we exploit it for the  action recognition context.}
\begin{table}[ht]
    \scriptsize	
    \centering
    \begin{tabular}{|c|c|c|}
        \hline
        \Black{Partioning 1} & \Black{Partioning 2} & \Black{\textbf{Ours}}\\
        \hline
        \begin{minipage}{.12\textwidth}
            \includegraphics[width=\linewidth]{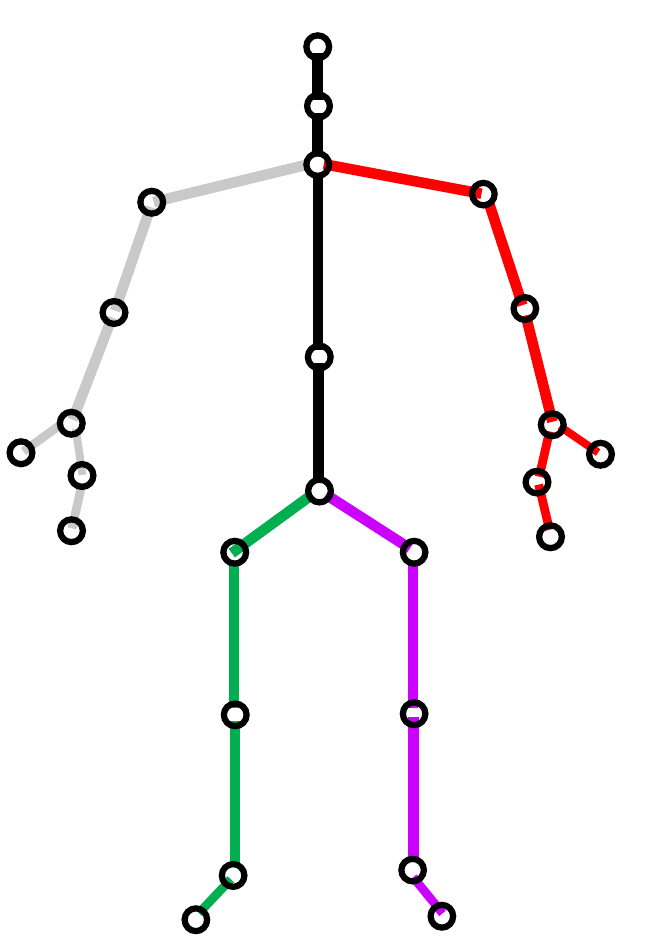}
        \end{minipage}
        & \begin{minipage}{.12\textwidth}
            \includegraphics[width=1.1\linewidth]{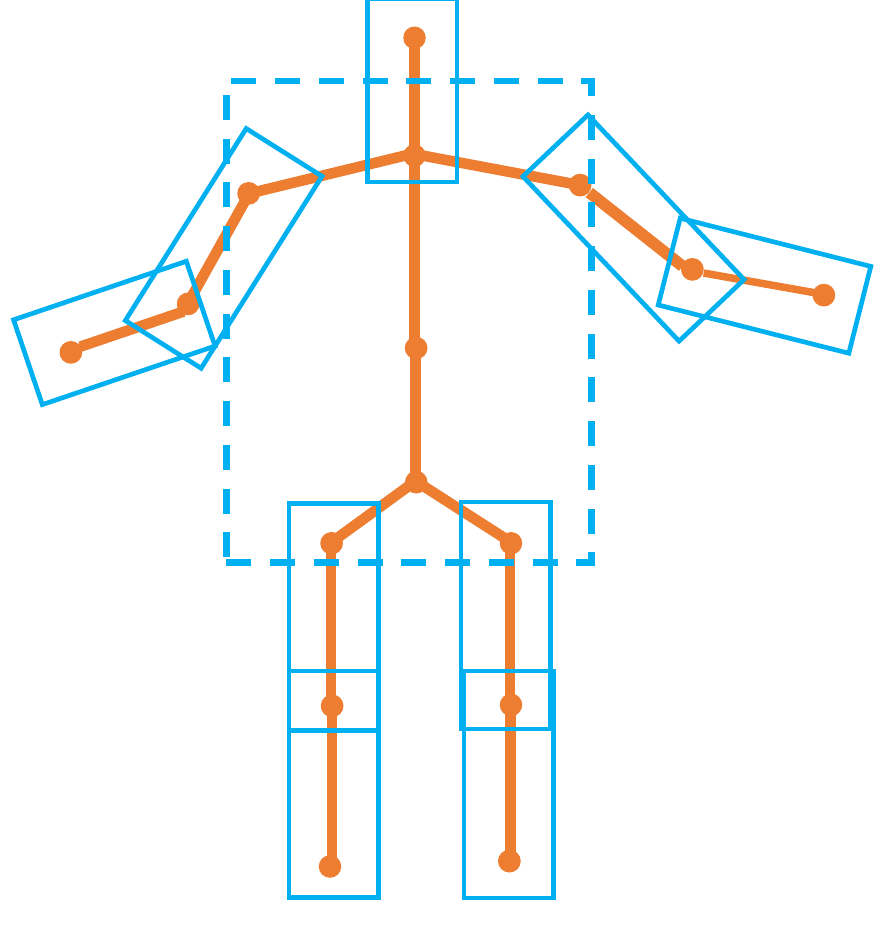}
            \end{minipage}
        &  \begin{minipage}{.12\textwidth}
            \includegraphics[width=\linewidth]{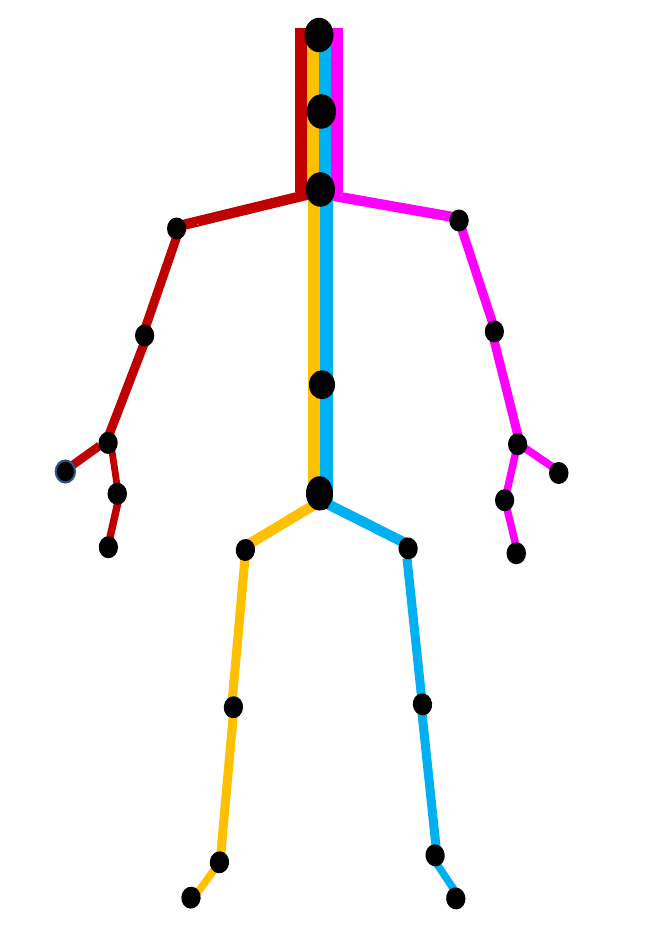}
        \end{minipage}\\
         \hline
         \Black{Acc: 94.46\%} &  \Black{Acc: 94.81\%} & \Black{Acc: 96.19\%} \\
         \hline
    \end{tabular}
    \caption{\Black{SPD Siamese network performance with respect to partitioning strategy on OAD segmented sequences. }}
    \label{partitioning}
\end{table}

\Black{\textbf{Classifiers performance:} 
Using the settings described in this section and the the body/hand partitioning mentioned in section ~\ref{PARTS}, we apply the SPD Siamese network on the different benchmarks: ODHG and \Black{SHREC 2021} for hand gesture recognition and OAD, UOW and InHard datasets for body action recognition. According to the Table ~\ref{classifier}, the performance of the proposed classifier are proved on both hand gesture and action recognition scenarios. In fact, an accuracy over 95\% is achieved in almost experimented datasets except for InHard  which is a new dataset very challenging one and is collected in real industrial environment.}
The achieved performances on action recognition datasets demonstrate that we have succeeded to generalize the SPD Siamese network on body skeleton sequences with an acceptable running time.
\begin{table}[ht]
\scriptsize
    \centering
    \caption{\label{classifier}SPD Siamese network performance on the datasets classifiers}
    \begin{tabular}{cccccc}
         Dataset & ODHG & \Black{SHREC} & OAD & UOW & InHard  \\
         \hline
          Accuracy(\%) & 95.60 & \Black{95.92} &96.19 & 98.84& 79.38  \\
          Running time(ms) & 173 & \Black{253} & 138 & 118& 352   \\
         \hline
    \end{tabular}
\end{table}

\subsection{\Black{Ablation study of the detector and online experiments}}

There are different levels of configurations to consider for the proposed detector: the one related to neural network and classification performance and the configuration of sliding window and verification process parameters.
The neural network of our detector, the SPD Siamese neural network, is kept with the same parameters set for the classifier.
Only interpolation step is considered with fewer number of frames since the detector training sequences has shorter.
Besides, many experiments are conducted to evaluate our system in term of running speed.
For the evaluation of the detection process, we use the following metrics: Jaccard index, prediction accuracy, prediction accuracy, F1-score, SL-score, EL-score. To better understand  these metrics, we refer readers to ~\cite{li2016online,caputo2021shrec}.

\begin{table*}[t]
    \scriptsize
    \centering
    \caption{Model performance for different settings of $ws$}
    \label{ws}
    \begin{tabular}{c|ccc|ccc|ccc|ccc}
    & \multicolumn{3}{c|}{ODHG} 
    & \multicolumn{3}{c|}{\Black{SHREC 2021}}
    &\multicolumn{3}{c|}{OAD} 
    & \multicolumn{3}{c}{UOW Online Action 3D}   \\
            \hline
        $ws$ (frames) 
        &12 & 18 & 24 
        & \Black{15} & \Black{20} & \Black{30} 
        & 4 &  6  &  10
        & 8  &  12  &  20   \\
        Detector R.T (ms) 
        & 93 & 100 & 105
        &52 & 58& 69 
        & 77 & 85 & 99.7
        & 80 & 82 & 85  \\
        
        \hline    
   
        Detector accuracy (\%)  
        & 73.11 & 75.89 &\textbf{ 77.12} 
        & \Black{ 92.76 } & \Black{92.82 } & \Black{\textbf{93.31}} 
        &86.97 &  88.34  &  \textbf{89.34} 
        &  61.13  &  76.09  & \textbf{ 77.5}    \\
        
        Detection  accuracy
        & 0.723 &\textbf{ 0.771} & 0.739
        & \Black{0.737 } & \Black{\textbf{0.770} } & \Black{0.615} 
        & 0.619 &   \textbf{0.901}  & 0.801 
        & 0.802  & \textbf{ 0.84}  &  0.812  \\
        
        SL-score  & 
        0.627 & \textbf{0.676} & 0.645
        & \Black{0.674 } & \Black{\textbf{0.694} } & \Black{0.544 } 
        &   0.561  &  \textbf{0.796} & 0.715 
        &  0.708  &  \textbf{0.747}  &  0.719   \\
        
        EL-score & 
        0.601 & \textbf{0.661 }& 0.632 
        & \Black{0.658 } & \Black{\textbf{0.691} } & \Black{ 0.526} 
        & 0.593& \textbf{0.803} &  0.715 
        &  0.73  & \textbf{ 0.733}  &  0.725  \\
        
        Gloabl F1-score  &
        0.713 & \textbf{0.769} & 0.732 
        & \Black{ 0.696 } & \Black{\textbf{0.720} } & \Black{ 0.632}
        &0.763 & \textbf{0.915}  &  0.859 
        &  0.852  & \textbf{ 0.881}  &  0.862 \\
        \hline
       
    \end{tabular}
    
\end{table*}

\begin{table*}[t]
    \scriptsize	
    \caption{ \label{te} Evolution of model performance according to $te$ }
    \centering
    \begin{tabular}{c|ccc|cc|ccc|ccc|c}
    &\multicolumn{3}{c|}{ODHG}
    &\multicolumn{2}{c|}{\Black{SHREC 2021}}
    &\multicolumn{3}{c|}{OAD}
    &\multicolumn{3}{c|}{UOW Online Action 3D}  
    & InHard \\
    \hline
    te 
    & 3 & 4 & 5 
    & \Black{3 } & \Black{5}
    &  2 &  3  &  5 
    & 2  &  3  &  5 
    & 3 \\
    
    \hline
    Detector accuracy  
    & \multicolumn{3}{c|}{75.89\%}
    & \multicolumn{2}{c|}{\Black{92.82\%}}
    & \multicolumn{3}{c|}{88.34\%} 
    &  \multicolumn{3}{c|}{76.09\%} 
    & 67.33\% \\
    \hline

    Detection  accuracy
    & \textbf{0.771} & 0.752   &0.742 
    &\Black{\textbf{0.770}}  & \Black{0.737}
    &  \textbf{0.901}  &  0.829    &  0.801 
    &  0.74  &  \textbf{0.84}  &  0.414  
    &0.675 \\
    
    SL-score  
    &\textbf{0.676}     &0.645  &0.63 
    &\Black{\textbf{0.694}} & \Black{0.665}
    &  \textbf{0.796}  &  0.762    &  0.727 
    & 0.675  &  \textbf{0.747}    &  0.351 
    & 0.578 \\
    
    EL-score  
    & \textbf{0.661}& 0.645        & 0.65 
    & \Black{\textbf{0.691} } & \Black{0.66}
    & \textbf{ 0.803}  &  0.775    &  0.762 
    &   0.664 &  \textbf{0.733}   &  0.341  
    &0.564 \\
    
    Gloabl F1-score  
    &\textbf{0.769 }   & 0.736   & 0.746 
    & \Black{\textbf{0.720} } & \Black{0.716}
    &  \textbf{0.915}  &  0.89    &  0.876 
    &  0.826  &  \textbf{0.881}    &  0.499  
    &0.661 \\

    \hline
    \end{tabular}
\end{table*}

\textbf{Influence of window size variation}

We recall that window size has to be long enough to detect the kinetic of human in continuous sequences but not too long  that it exceeds some human action duration or affects the detector rapidity.
The kinetic state needs to be detected 4 or 5 times per second at least. We set $r=$  { 6, 5, 2, 4, 6} frames to be the refresh rates respectively of ODHG, SHREC 2021, OAD, UOW and InHard datasets.
Then, we vary $ws$ values corresponding to each dataset (see Table~\ref{ws}).
The different variations of $ws$ regarding each dataset is due to the difference of the sensors capturing rate.
\Black{
According to the results described in Table ~\ref{ws}, the detector running time did not exceed 120 ms even when varying $ws$ in different intervals. It enables us to work online and detect the kinetic state 4 or 5 times per second, which is an acceptable frequency on human action recognition task.}
\\
Regarding the performance of the detector, we remark that the accuracy of the detector is enhanced by the elevation of the window size in the datasets. This is expected since with this elevation, the detector is able to generate more information about the motion evolution. However, the behavior of the other metrics is different, as they continue to rise until a peak value $\hat{ws}$ is reached. ($\hat{ws} =${18, 20, 6, 12, 18} are respectively the peak values of window sizes for ODHG, SHREC 2021, OAD, UOW and InHard datasets). For a longer window size, the reported performances shrink.
This can be explained by to the fact that the exaggerated increase of  windows size makes the window covering rapidly different transitions of kinetic status in the case of actions with short duration. This causes ambiguity in identifying the true kinetic state and affects the system performance, especially in motion boundaries prediction (SL-score and EL-score remarkably decrease).
Reducing the number of frames excessively will lead to a lack of available information about the ongoing state for the detector and thus will lead to difficulty in the online recognition.
\Black{It is remarkable that the best performances correspond to a window size compromise between 0.6s and 0.8s. This Time interval can be considered to be the interval of visibility for the detector.}

\textbf{Influence of the V.P parameters variation}

In this experiment, we vary $te$: the number of tests performed by the detector  during the verification process.
The results are shown in the Table ~\ref{te}.
We remark that the best performance is given by $te=3$ for all the datasets except OAD for which the peak is given by $te = 2$. This difference is due to different reasons. Firstly, the detector performance of OAD dataset is the highest performance comparing to the other detectors. Since the tests are performed in order to reduce the error rate at the detector level, it is expected that the detector of the higher performance will need a smaller number of tests. More importantly, the actions durations are arbitrary in OAD dataset: some actions last less than half a second and others last more than half a minute. The detector extends the window size to 10 frames (1.25s) and may reach the idle period while performing 3 tests. 
In contrast, motions in other datasets example are more regular and without stagnation. Working with $te > 5$ gives performances slightly inferior to those reported with $te=3$ and with more time. So, we prefer to not exceed 3 tests.
\subsection{Early classification experiments}
\begin{figure}[ht]
    \centering
    \includegraphics[width=0.466\textwidth]{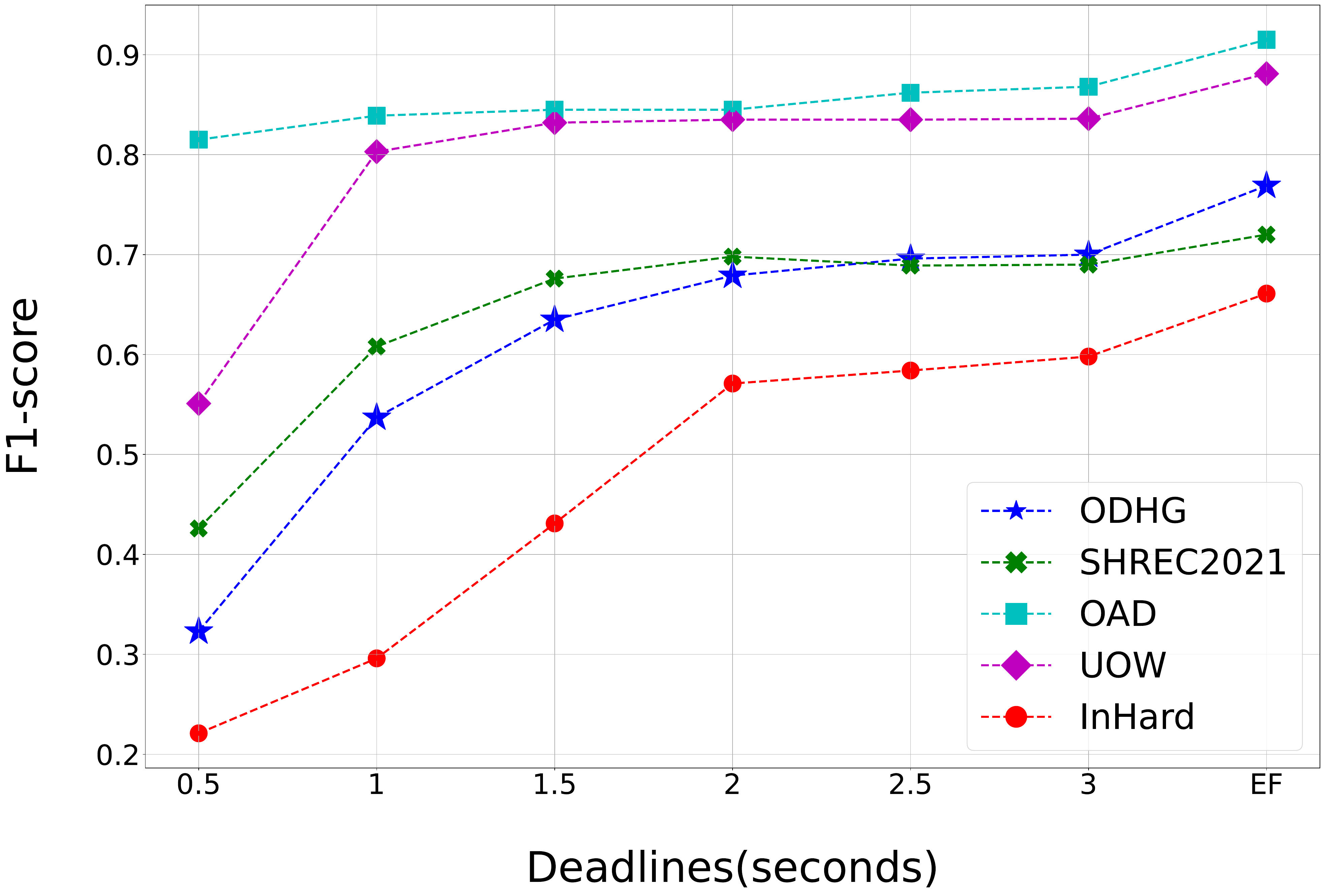}
    \caption{Model behavior toward early classification different deadlines.}
    \label{early-class}
\end{figure}

We set different deadlines for classification: $T$= [0.5s, 1s...3s]. We evaluate the performance of our system and to what extent it can maintain its efficiency. The obtained F1-scores are described in Figure ~\ref{early-class}.

Observing the obtained curves, we remark that the model didn't show an efficiency in the lower deadlines. This can be explained that human activities generally need more than a second to be completely performed in the most cases. Higher than one second, the curves show F1-scores amelioration especially for OAD, UOW and \Black{SHREC 2021} which approach rapidly to the ideal case. For ODHG and InHard, it needs more time to move towards the F1-score of the no-early classification case because the performances of their detectors are lower than those of the other datasets. Also, they have some motions which are performed in the same manner at the beginning. For example, the first part of gesture "Tap" and gesture "Swipe Down" in ODHG dataset are too similar.

\subsection{Comparison with the state-of-the art}
In order to assert the efficiency of our system, we need to compare its performance with the previous proposed approaches.
\Black{Our experiments are reported on both online hand gesture and online action recognition contexts.}

\Black{\textbf{ Comparison of Running Speed:}}

\Black{
Experiments on answer time are carried out on SHREC 2021 challenge ~\cite{caputo2021shrec} and comparison is made with different approaches proposed by four groups who give their performances on this dataset.
The average running time taken by the detector to identify the kinetic state in each window and in the whole sequences are described in Table ~\ref{R.T}.
}

\begin{table}[ht]
\scriptsize	
    \centering
    {\color{black}\begin{tabular}{c|c|c}
        Group &Class time (s)& Total time (s)\\
        \hline
        $1  $ ~\cite{caputo2020sfinge,caputo2021shrec} & $   1.36    $& $   435.5$ \\
        \hline
        \multirow{1}{*}{2}~\cite{vaswani2017attention,caputo2021shrec}&  $0.41$ &  $   48781.0$ \\
        \hline
        \multirow{1}{*}{3} ~\cite{maghoumi2020deep,caputo2021shrec}&  $   \textbf{0.6}\times\textbf{10}^{\textbf{-4}}$ & $  \textbf{ 66.7} $ \\
        \hline
        \multirow{1}{*}{4} ~\cite{yan2018spatial,caputo2021shrec}&  $0.066$ & $  94.6 $ \\
        \hline
        $Ours $ &  $ 0.038 $ & $ 3836.88$ \\
        \hline
        
    \end{tabular}}
    \caption{\Black{Execution time corresponding to different approaches on SHREC 2021}}
    \label{R.T}
\end{table}
\Black{We remark that the average running times per single prediction (class time) are very discarded. Our system takes about 69 ms to give a single prediction, enabling it to identify the kinetic states of more than 12 windows per second in average. This frequency is acceptable in real live applications. It is faster than approaches used by groups 1, 2 and group 4. However, group 3 proposed a faster approach, thanks to the simple complexity of its model which helps for better responding time but not highly efficient in term of performance ~\ref{sota1}.}

\Black{For the global classification, our system takes about 3837 seconds to compute all the test sequences. These sequences contain more than 119K frames, capturing at 25 fps. They last about 4700s, which is close to the total time taken by our system because the refresh rate of the system is designed to make the system execution in phase with real streaming.}

\Black{\textbf{State-of-the-art on hand gesture context datasets:}}

\Black{For hand gesture recognition, we evaluate and compare the performance of our online system on  SHREC 2021  dataset ~\cite{caputo2021shrec}.
Best performances of research groups participating in SHREC 2021 challenge are reported in Table ~\ref{sota1}. Only skeletal based approaches are considered for this comparison.
The proposed metrics for evaluation in this challenge are Detection rate, Jaccard index and FP rate.}
\begin{table}[ht]
    \scriptsize	
    \centering
    {\color{black}\begin{tabular}{M{46.9mm}| M{9.3mm}| M{7.5mm} | M{5.5mm}}
        Method & Detection rate & Jaccard index& FP rate\\
        \hline
        Dissimilarity-based Classification ~\cite{caputo2020sfinge,caputo2021shrec} & 0.3993 & 0.2566 &0.764\\
        Transformer module ~\cite{vaswani2017attention,caputo2021shrec} & 0.7292 & 0.6029 &0.257\\
        uDeepGRU+TSGR ~\cite{maghoumi2020deep,caputo2021shrec}& 0.7431 & 0.6238 &0.271\\

        ST-GCN ~\cite{yan2018spatial,caputo2021shrec}& \textbf{0.8993} & \textbf{0.8526} & \textbf{0.066}\\
        \hline
        Ours & 0.770 & 0.667 & 0.230\\
        \hline
    \end{tabular}}
    \caption{\Black{Performance comparison with the state-of-the-art online methods on SHREC 2021 gesture benchmark dataset}}
    \label{sota1}
\end{table}

\Black{According to Table ~\ref{sota1}, our system reveals interesting performance and shows its superiority over the majority of the proposed models. Only adapted ST-GCN approach seems to outperform our system performance. 
This can be explained by the fact that the used approach is not only based on online ST-GCN but uses also Trajectory-based fine tuning approach to simplify the recognition in some specific gestures. Besides, energy-based detection module seems to be a good choice on such context.}

\Black{\textbf{State-of-the-art on body action context datasets:}}

\Black{Comparison with state-of-the-art approaches on online action recognition datasets regarding the commonly used metrics F1-score and prediction accuracy  are reported in Table ~\ref{sota2}.
We can notice that our model outperforms  other models, especially in  UOW Online Action 3D dataset, which witnessed an increase of 17\%. Besides, \Black{we achieve the best performance on OAD dataset in term of F1-score and prediction accuracy.} This proves the efficiency of the detector, the key role played by the SPD Siamese model in ensuring the quality of the results given by the system and the importance of the tests realized in the beginning and the end of each action. Finally, promising results are highlighted in the challenging InHard dataset which represents specific industiral context. In fact, actions in continuous sequences are performed without stagnation periods and with human-robot collaboration scenarios in some cases.}

\begin{table}[t]
    \scriptsize	
    \centering
    \caption{Performance comparison with the state-of-the-art methods on body action context datasets}
    \begin{tabular}{M{22.5mm}| M{5mm} M{5mm}| M{5mm} M{5mm}| M{5mm}M{5mm}}
    Dataset & \multicolumn{2}{c|}{OAD} & \multicolumn{2}{c|}{UOW} & \multicolumn{2}{c}{InHard}\\
    \hline
         \diagbox{Method}{Metric}& F1-score& \Black{Pred. Acc}&F1-score& \Black{Pred. Acc}&F1-score&\Black{Pred. Acc}\\
         \hline
         RNN-SW
       ~\cite{zhu2016co}&0.600 &-&-&-&-&-\\
         JCR-RNN
       ~\cite{li2016online}&0.653 &\Black{0.788} &-&-&-&-\\
        \Black{ST-LSTM} ~\cite{liu2017skeleton}&-&\Black{0.770}&-&-&-&-\\
        \Black{Attention Net} ~\cite{liu2017global} &-&\Black{0.783}&-&-&-&-\\
         RF+ST
       ~\cite{baek2017real}&0.672 &-&-&-&-&-\\
         
         MM-MT-RNN
       ~\cite{liu2018multi}&0.795 &-&-&-&-&-\\
        \Black{FSNet} ~\cite{liu2019skeleton}& -&  \Black{0.800}&- &- & &-\\
         \Black{SSNet} ~\cite{liu2019skeleton}& -&  \Black{0.820}& -&- & &-\\
         \Black{VGG16+L1} ~\cite{mokhtari2022human} &-&\Black{0.868}&-&- &-&-\\
         SW-CNN
       ~\cite{delamare2021graph}& -& -& 0.680&\Black{0.680}&-&-\\
         SW-GCN
       ~\cite{delamare2021graph}& -&-& 0.750&\Black{0.755}&-&-\\
         \hline
         Our model & \textbf{0.915} &  \Black{\textbf{0.901}}& \textbf{0.881}& \Black{\textbf{0.840}} &\textbf{0.661}&\Black{\textbf{0.631}}\\
         \hline
    \end{tabular}
    
    \label{sota2}
\end{table}

\section{Conclusion}
We proposed an accurate  manifold-based approach which works either for hand gesture sequences or for human body action recognition. We built an online system composed from a detector and a classifier. The detector is used for the prediction of the action intervals using a verification process and the sliding window approach. The classifier uses the segmentation predicted by the detector for motion recognition. We provided the experimental evaluation on five benchmark datasets and compared our approach with the recent state of the art methods. However, our approach can be more optimized, especially the classifier architecture which can help to obtain lower execution time.
As future work, we plan to study more the human motion recognition challenges especially in industrial context such as human robot interaction in real case study and in virtual environment.

\bibliographystyle{unsrt}

\bibliography{bibliography.bib}

\end{document}